\documentclass[sigconf, screen]{acmart}
\AtBeginDocument{%
  \providecommand\BibTeX{{%
    \normalfont B\kern-0.5em{\scshape i\kern-0.25em b}\kern-0.8em\TeX}}}

\setcopyright{acmcopyright}
\copyrightyear{2021}
\acmYear{2021}
\acmDOI{no doi}

\acmConference[HRI '21]{HRI '21: Exploring Applications for Autonomous Non-Verbal Human-Robot Interactions
at the ACM/IEEE International Conference on Human Robot Interactions}{March 08--12, 2021}{Virtual Conference}
\acmBooktitle{HRI '21: ACM/IEEE International Conference on Human Robot Interactions}
\acmPrice{}
\acmISBN{}

\begin{document}

\title{An explorative study on how human-robot interaction is taken into account by robot developers in praxis}

\author{Doris Aschenbrenner}
\email{d.aschenbrenner@tudelft.nl}
\orcid{0000-0002-3381-1673}
\affiliation{%
  \institution{TU Delft IDE}
  \streetaddress{Landbergstraat 15}
  \city{Delft}
  \country{Netherlands}
  \postcode{2628 CE}
}

\author{Danielle van Tol}
\affiliation{%
  \institution{TU Delft IDE}
  \city{Delft}
  \country{Netherlands}}
\email{D.H.vanTol@tudelft.nl}

\author{Pak Long Cheung}
\affiliation{%
  \institution{TU Delft IDE}
  \city{Delft}
  \country{Netherlands}}
\email{P.L.Cheung@tudelft.nl}

\author{Zoltan Rusak}
\affiliation{%
  \institution{TU Delft IDE}
  \city{Delft}
  \country{Netherlands}}
\email{Z.Rusak@tudelft.nl}

\renewcommand{\shortauthors}{Trovato and Tobin, et al.}

\begin{abstract}
How is human-robot interaction considered within the development of new robotic systems by practitioners? This study sets out to inquire, whether the development teams of robotic products have been considering human factor methods in their design and implementation process. We were specifically interested in the non-verbal communication methods they were aiming to implement, and how they have approached the design process for these. Although valuable insights on tasks and communication needs during the different phases of robot operation could be gathered, the results of this study indicate, that the perspective of the human user or bystander is very often neglected and that knowledge on methods for engineering human-robot interaction is missing. The study was conducted with eleven development teams consisting of robot manufacturers and students within a robot building course representing overall 68 individual participants.  
\end{abstract}




\maketitle

\section{Introduction}
Robotics research was for a long time majorly performed in the narrow domains that could afford the high investments necessary to build capable robotic systems \cite{goodrich2008human}. Thus, robots have been mainly applied for areas like space exploration or within other hazardous environments, for example, search and rescue applications. Here it was clear, that humans cannot operate on their own and need some kind of mechanic assistant. Furthermore, industrial robots have been developed for the production environment, in which they take over automation tasks and were kept separate from the workers behind safety fences. As robotics research has advanced, the field of “social robots” for hospital or home environments has been opened. Here, the necessity to take human safety and interaction into account is logical and have immersed into an own field of study. Within the increasing amount of discussions on the “future of work” \cite{lorenz2015man}, it becomes clear that there is a possible future in which humans might work alongside intelligent machines and robots in their daily work routine in a “hybrid intelligence” \cite{dellermann2019hybrid} setting. But are the current systems designed so that they can meet this goal? If they are not (yet), are at least the development teams aware of the implications of human-centered approaches within robotic system development, so this goal can be achieved? As already pointed out by Sheridan \cite{sheridan2016human}, there is still a principle gap between the human factors professionals and the robotics engineers. Specifically in one of the “origin” fields of robotics research, the production industry, there are many studies \cite{ludwig2018revive} \cite{romero2016operator} that advocate for looking at the entire socio-technical system and thus strengthening human factor aspects. The consideration of human-robot interaction principles within the design of robotic systems is highly relevant for guaranteeing i) high performance and ii) high well-being of the involved humans. Thus, the explorative study presented in this publication aims at finding out the following research questions:
\begin{itemize}
    \item How do current robotics development teams approach the design challenge in general?
    \item Is there interaction with the target user group during development?
    \item What do members of robotic development teams think of the user of their system?
    \item Which methods from human factor research are used?
    \item Which kind of (verbal and non-verbal) interaction do they envision for their product?
\end{itemize}

\section{Approach based on existing taxonomies}

\begin{figure}[h]
  \centering
  \includegraphics[width=0.8\linewidth]{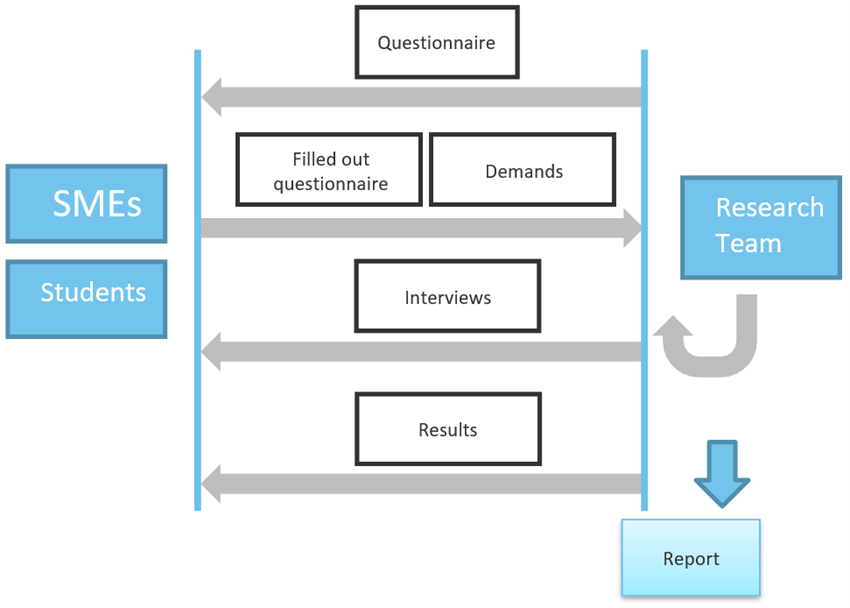}
  \caption{Approach of this research}
  \label{fig:approach}
\end{figure}

The research was conducted in two steps, which is displayed in Figure \ref{fig:approach}: First, a questionnaire has been developed and sent out to companies and students of a robotic development university course. Then, a round of interviews was performed on the same target group.

\begin{figure}[h]
  \centering
  \includegraphics[width=0.8\linewidth]{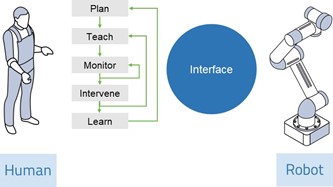}
  \caption{Human robot interaction classification according to task model from Sheridan \cite{sheridan1992telerobotics}}
  \Description{Human tasks Plan, Teach, Monitor, Intervene, Learn}
  \label{fig:taskmodel}
\end{figure}

\begin{figure}[h]
  \centering
  \includegraphics[width=0.7\linewidth]{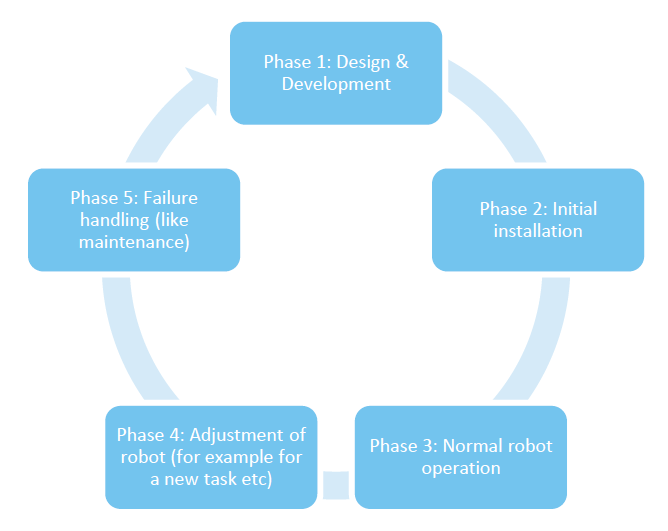}
  \caption{Lifecycle phase model for robot development}
  \Description{Lifecycle Phase 1: Design & Development, Phase 2: Initial installation, Phase 3,: Normal robot operation, Phase 4. Adjustment of robot, Phase 5: Failure handling)}
  \label{fig:lifecycle}
\end{figure}

The first aspect was used to inquire on how and if the investigated methods fit into the existing taxonomies discussed in the literature. We decided to inquire basically into the theoretical models displayed in Figure \ref{fig:taskmodel}, the revised taxonomy of human-robot interaction from Yanco and Drury \cite{yanco2004classifying}, and a self-developed phase model in Figure \ref{fig:lifecycle} which describes a generic robot lifecycle.

\begin{figure}[h]
  \centering
  \includegraphics[width=\linewidth]{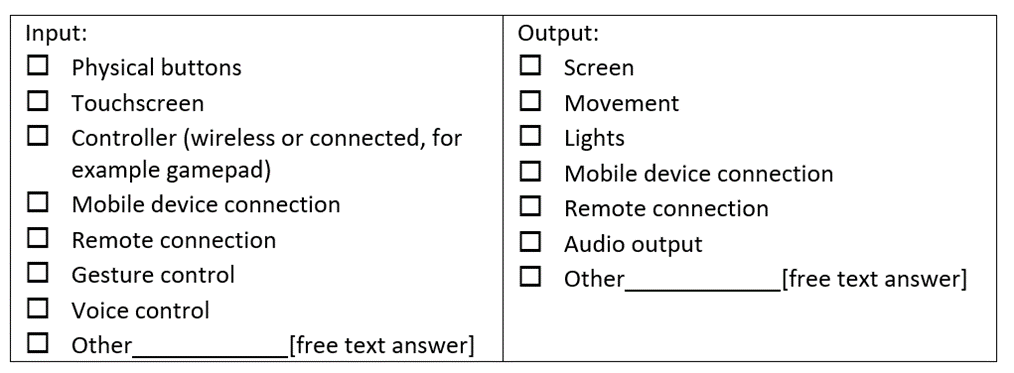}
  \caption{Communication input and output}
  \label{fig:nonverbal}
\end{figure}

Concerning non-verbal communication, we used the questions displayed in Figure \ref{fig:nonverbal} to investigate the “means of interaction” that have been used within the participant teams.

\section{Questionnaire study}

\subsection{Experimental Design}

The experiment was designed as a questionnaire with subjective reporting and can yield both qualitative and quantitative results. The questionnaire has been sent over a mailing list with robotic developers (small and medium-sized companies, SMEs) that have received funding through a European cascade funding and was distributed to an interdisciplinary course within TU Delft, the “Robotics minor”, in which groups of students from four to five faculties build a robot system from scratch together following an assignment of a paying client from outside the university. The initial idea was to compare the students with professional teams, as the student has gained basic training on human factor methods, and with the professional teams, the status of existing knowledge is unknown. Due to the low reply rate, a statistically significant comparison was not feasible.

\subsection{Experimental Procedure}

The participants have been asked to fill out the questionnaire which will be made available under \url{http://smartfactoryresearch.com}. The questionnaire consists of four parts:

\begin{enumerate}
    \item \textbf{Your project and team} – statistical data on team composition and project description
    \item \textbf{The target user in your project} – in order to identify and characterize user groups following a self-developed phase model, combined with the task model from Sheridan and the revised level of interaction taxonomy (all discussed above)
    \item \textbf{Human-Robot Interaction in your project} – based on the identified user groups, this part of the questionnaire aims at inquiring on the existing knowledge of the user and their interaction and communication methods
    \item \textbf{User interaction research/ design in your project} – this part inquires further on the applied human factor methods and the existing knowledge of these approaches
\end{enumerate}

\subsection{Results}

\subsubsection{Participants}
It was very complicated to receive feedback from companies, after numerous attempts we merely were able to get four filled-out questionnaires. Together with the student groups, we were able to analyze eleven feedback questionnaires. In total, there are 68 people involved in the development teams, the mean amount of people involved per group is 6.2 where the smallest groups reported three team members and the largest group consists of sixteen members.
\textit{Gender balance}: In the SME group we found two teams with equally distributed gender and two male-dominated teams. In the student group, there were two teams with only men, the others were male-dominated.
\textit{Discipline balance}: Both groups consist mainly of software-engineers (19 team members from 68) and the user experience designer underrepresented (1 from 68 team members).
\textit{Age balance}: Both groups were constituted by young teams (age under 30), which also was the case for the answering company teams.
 
\subsubsection{Main user}
Based on the lifecycle phase model from Figure \ref{fig:lifecycle}, we will present a list of user groups that have been mentioned the most. 
\textit{Phase 1 (Design \& Development)}: “development team”, “software engineers”, “us”, or “\$companyname staff”, whereas \$companyname is used here as a placeholder for the different mentioned company names.
\textit{Phase 2 (Initial installation)}: “development team”, “software engineers”, “us”, or “\$companyname staff”.
\textit{Phase 3 (Normal robot operation)}: The main user mentioned were unspecific like “client”, “robot operator”, or “\$companyname employees”, but some were more specific like “medical patients” or “assembly line workers”.
\textit{Phase 4 (Adjustment of robot)}: “software engineers”, “robotics development staff”, “tech support worker”, “development team client”.
\textit{Phase 5 (Failure handling)}: “tech support workers”, “Mechanical engineers at \$company”, or  / “\$company maintenance technicians”.

\subsubsection{Identified human tasks} 
\begin{figure}[h]
  \centering
  \includegraphics[width=\linewidth]{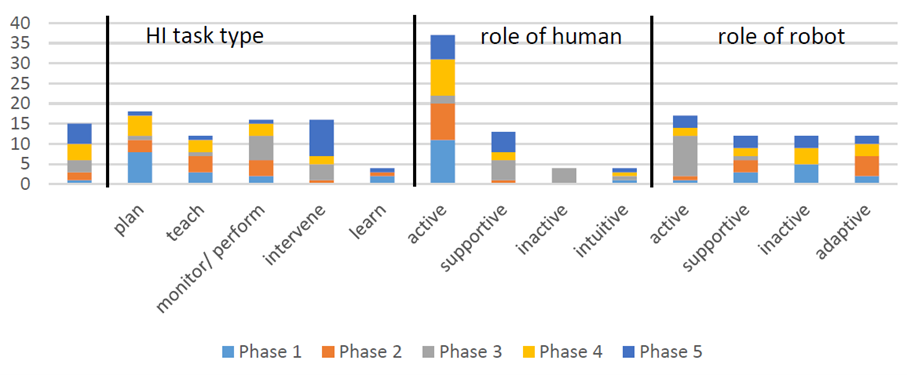}
  \caption{Analysis results on task type (Sheridan) and interaction type (Yanco and Drury) with respect to the different phases of the robot lifecycle. $y$ is the absolute number of mentions}
  \label{fig:humantasks}
\end{figure}

In Figure \ref{fig:humantasks} the summary of the identified tasks of the human (Sheridan) are combined with the reported interaction role of human and robot (Yanco and Drury). Interesting is, that the human seems to be active in all phases, but least in the third phase (normal operation). Only in this phase, the human has been considered “inactive” by some participants. The robot is mainly active in the third phase, which is logical, but interesting enough the robot also should take a role in the other phases. The user appears to mainly “Plan” in phase 1, mainly “Teach” in Phase 2, “Monitor” in Phase 3, and “Intervene” in Phase 5.

\subsubsection{Means of interaction} 
\begin{figure}[h]
  \centering
  \includegraphics[width=0.65\linewidth]{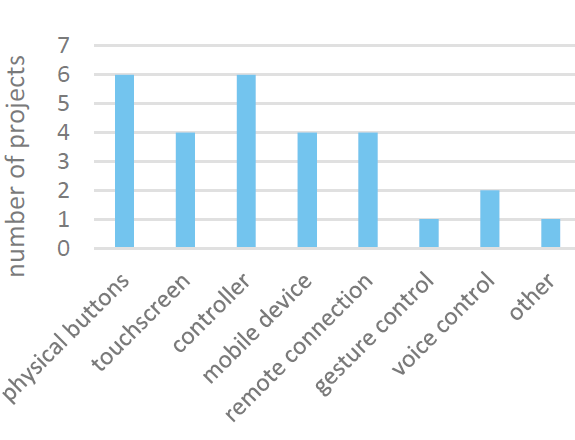}
  \caption{Means of interaction mentioned by project feedback (multiple answers have been possible) for input. $y$ is the absolute number of mentions}
  \label{fig:input}
\end{figure}

\begin{figure}[h]
  \centering
  \includegraphics[width=0.65\linewidth]{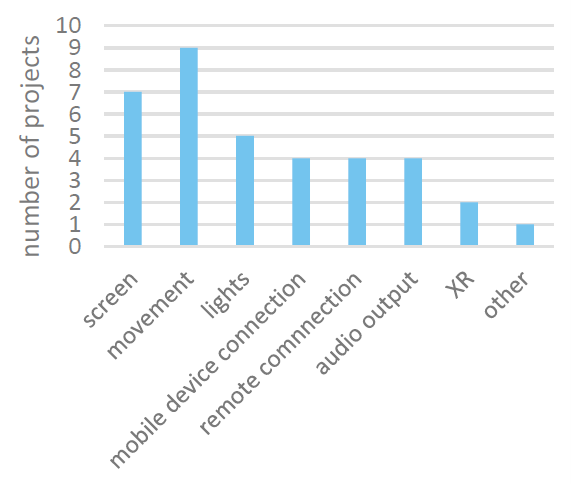}
  \caption{Means of interaction mentioned by project feedback (multiple answers have been possible) for output. $y$ is the absolute number of mentions}
  \label{fig:output}
\end{figure}

Figure \ref{fig:input} provides an overview of the mentioned input variants and Figure \ref{fig:output} presents the output means. The main mentioned interaction for the Input were physical buttons (6) and controller (6), and as the main output means the robot’s movement (9) and the screen (7) has been mentioned.

\subsubsection{Communication needs} 
\begin{figure}[h]
  \centering
  \includegraphics[width=\linewidth]{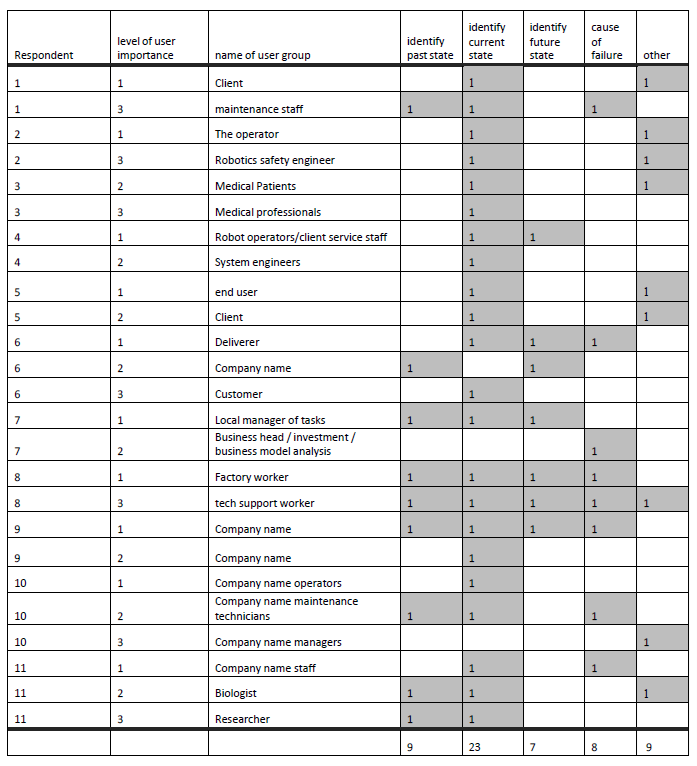}
  \caption{Communication needs from the human side}
  \label{fig:communicationneeds}
\end{figure}

As mentioned above, the questionnaire asks the participant to mention their most important user per phase and then extract their three main important user groups. We wanted to find out the communication needs and the required human skills by asking a question on the tasks that the target user groups are performing with the robotic system. We decided to group the response options according to situation awareness / legibility, add failure identification, and leave an open field for free annotations (other). The results are displayed in Figure \ref{fig:communicationneeds}. Interestingly enough (but maybe motivated by the questionnaire design), the own development team was mentioned six times as being part of the three main important user groups, this is omitted in Figure \ref{fig:communicationneeds}. Furthermore, empty entries have been removed. Only one group mentioned themselves as the user with the highest priority, which appears a strange thing as they have been clearly developing a robot that was to be used not by themselves. The most relevant information need seems to be to identify the current state. The feedback to the “other” category was partially quite verbose, mentioning various different tasks that the users should be able to fulfill. Some examples on the replies for the “other” tasks are: “Benefit from the assistive applications of the robot”, “Teach the task to the robot”, “Robot does medical procedures on this group”, "Manually control the robot to move around a beach.”, “Identify the current state of the robot (location, active/inactive)”, “Create a map that contains the locations of cigarette butts in the beach.”. Furthermore, we asked an additional question specifically about the skills that the operators need to have. It is interesting that, according to the responses, the end-users mainly need to have “no” skills within phase 3, and already, for phase 4, possess relatively high knowledge (like “robotic development”), although minor adjustments will be part of every robot routine.

\subsubsection{Applied human factor methods} 
\begin{figure}[h]
  \centering
  \includegraphics[width=\linewidth]{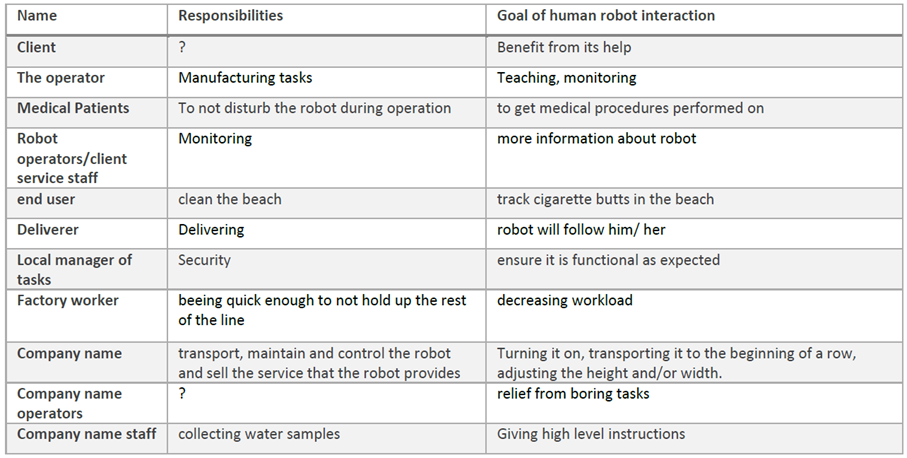}
  \caption{Goal of human robot interaction}
  \label{fig:goalHRI}
\end{figure}

\begin{figure}[h]
  \centering
  \includegraphics[width=\linewidth]{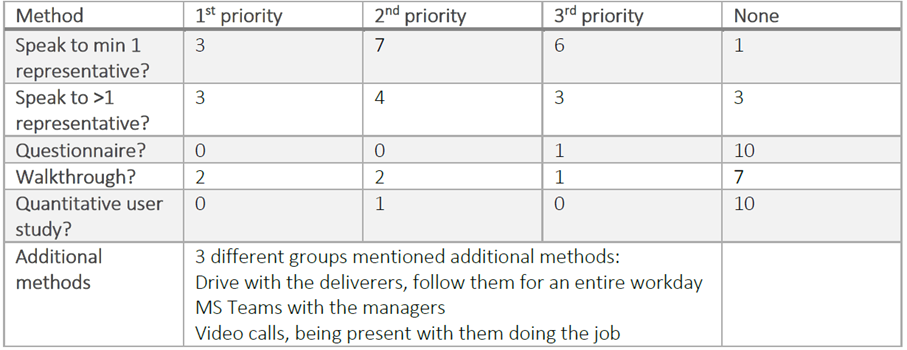}
  \caption{Human factor methods applied so far (please note, that the six times mentioned own team as a target user group is part of the presented data)}
  \label{fig:HFmethods}
\end{figure}

Next to this, we inquired upon the responsibilities of the highest important user group and the goal of the human-robot interaction and displayed the results in Figure \ref{fig:goalHRI}. We wanted to know, which human factor methods have already been applied within the interaction with the identified most important target user groups (Figure \ref{fig:HFmethods}). Five teams answered the questions on whether there are any bystanders with “yes, but we do not take them into account”, four groups answered with “Yes, we take them into account”, and two groups answered with “No, there are no bystanders”. Furthermore, we asked, whether there already have been any interaction problems identified. Here, six groups answered “No”, which, from an HRI research perspective, might be due to the fact that there has not been enough interaction with the target user group yet. Additionally, we asked whether they think they need to know more about their users. Here, four groups answered with “No”, and seven groups answered with “Yes”. We also asked whether the group, after finishing the questionnaire, would like to know more about their users. Here, only one group answered with “No” and the other ten answered with “Yes”.

\section{Interviews study}

\subsection{Experimental Design}
\begin{figure}[h]
  \centering
  \includegraphics[width=\linewidth]{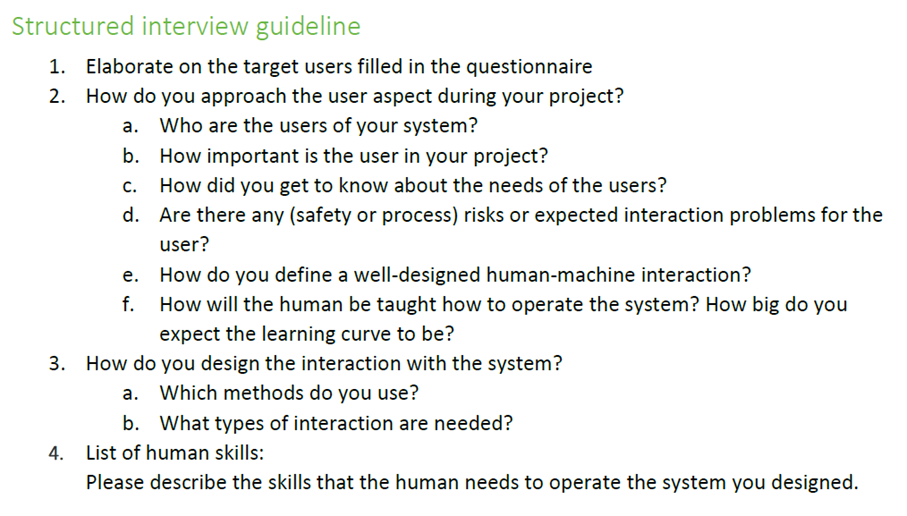}
  \caption{Structured interview guideline}
  \label{fig:interviewguide}
\end{figure}

We prepared a semi-structured interview, the leading questions can be found in Figure \ref{fig:interviewguide}. They consist of four blocks: First, we wanted to know more about their target users, after which we discussed how they approached the target users, then we asked about the design of the interaction, and finally on the skills that the human needs to operate the system.

\subsection{Experimental Procedure}

Unfortunately, none of the SME participants were interested in an interview, so this study is limited to the seven student teams. Within the education program, there have been interviews with the entire team of each of the seven student groups. The interviews took between 15 and 25 minutes and were conducted via Zoom and recorded. Afterward, the discussion was summarized in writing. The measured data were qualitative self-reporting on estimations and beliefs. It was tested how much the developer teams know about their user group and which methodology they applied.

\subsection{Results}
We present a summary of the findings in the following list:
\begin{itemize}
\item Not every client puts importance on the user of the robot. When the client does not specifically request good usability, an engineering team does not automatically think about that themselves. Also, the client is in some projects seen as the end-user with a strong opinion on the usage without actually being the person who is going to use it.
\item All groups want the robot to be intuitive and think that the user does not need any specific skills, but when delving deeper into it, there are always ways of interaction and communication that the user has to understand or learn.
\item In the view of the some developers, the projects primarily had the goal of a proof of concept on the technological aspect. These types of projects tend not to take the user into account, but the user aspect might still have a large impact on the feasibility of the system (if it works but no one wants to use it, then it cannot be successful).
\item It is important to have some sort of illustration or storyboard that illustrates the envisioned usage so that everyone sees it the same way and that all points of interaction are taken into account.
\end{itemize}

Specifically asking on communication and skills for the humans, the following quotes have been gathered:
\begin{itemize}
    \item “They don’t need any additional skills, although some training or ‘play time’ will help the delivery workers to better grasp the movement of the robot and how it steers.”
    \item “The operator of the robot does not need additional skills. Although a little bit of understanding of how the robot moves can be beneficial for extra safety.”
    \item “For the people that services the robot, there should be a guide so they know which panels to remove so you can reach the needed parts. We also want to incorporate error codes and intuitive language (if the robot flashes red light, something is wrong)”
    \item “Not much. Some basic controller skills should be enough. (it’s as simple as getting it out, putting two poles in the ground, drive it to the general direction and let it do its thing).”
    \item “When something goes wrong, an error will be displayed on the screen and the robot will shut down along with a description of how to solve it on the UI (e.g. ‘we detect an extreme angle, please put the robot on a flat surface). If the error persists after restarting, the UI will display ‘Contact engineers’.”
    \item “Apart from the specific setup (put in empty capsules and take out full capsules) and some instructions on how the robot is supposed to function, what you should do to prepare it, and what tasks need to be performed to make it ready for the next use.”
\end{itemize}

\section{Summary and discussion}

\subsection{Research questions}
First of all, the initially mentioned research questions should be answered based on the knowledge acquired from the questionnaire and interviews:
\subsubsection{How do current robotics development teams approach the design challenge in general?}
Based on the received feedback, it can be found, that most teams consider their robot as fully autonomous because it does not rely on (permanent) human input. Although they might succeed in the more technical aspects of the design challenge, they do not consider that human-robot interaction is a necessary field that needs to be regarded.
This can be supported by the fact, according to the feedback, the robot should be “active” during "normal operation" of the robot (Phase 3), whereas the human should mainly be "supportive" or "inactive" (Figure 19). It can also be supported by the written feedback we got from people who were directly asked by the research team to fill out the questionnaire – either they considered themselves to be merely component developers, did not have time, stated that their robotic system is “too autonomous” and thus the questions would not apply, or did not respond at all.
Interestingly, the results from the questionnaires but also the interviews show, that it is possible for all groups to identify a relevant user after asking several consequent questions, as the adjustment phase (Phase 4) requires attention from a human team which appears to be different from the maintenance and troubleshooting state (Phase 5), who are no experts in robotics. For these target groups the interaction with the robot should be “easy” and “intuitive”, as it can be derived from the interviews, so this is a relevant design challenge.
\subsubsection{Is there interaction with the target user group during development?}
Only three groups spoke with one or more representatives of the target user group with the highest priority (which includes the group which had positioned themselves as the most important user group). Thus it can be concluded, that it is very little or no interaction between the development team and the target user group. Especially, as after filling out the questionnaire, seven out of eleven groups had the feeling that they need to know more about their target user and ten out of eleven groups said, that it would improve the result of their project if they knew more about the target user.
\subsubsection{What do members of robotic development teams think of the user of their system?}
From the results of the research, it can be concluded that the development teams as a majority, do not care about the end-user. The most representative quote for this mindset was the feedback from a company team, which stated that the user's responsibilities were “To not disturb the robot during operation” (Figure \ref{fig:goalHRI}). As the specific application is the performance of medical procedures without another human person present, the interaction design is way more important than this specific answer implies. 

\subsubsection{Which methods from human factor research are used?}
This is answered in the findings of Figure \ref{fig:HFmethods}: There have been no advanced methods used, like a questionnaire or quantitative user studies, in order to investigate further into the human factors components of the highest priority user.
Two groups used a walkthrough method, but the majority did not. Additional methods have been mentioned which are rather applied in the contextual inquiry phase of a project but not during development. This might also explain the finding that the majority of the teams did not encounter interaction problems.
\subsubsection{Which kind of (verbal and non-verbal) interaction do they envision for their product?}
\begin{figure}[h]
  \centering
  \includegraphics[width=\linewidth]{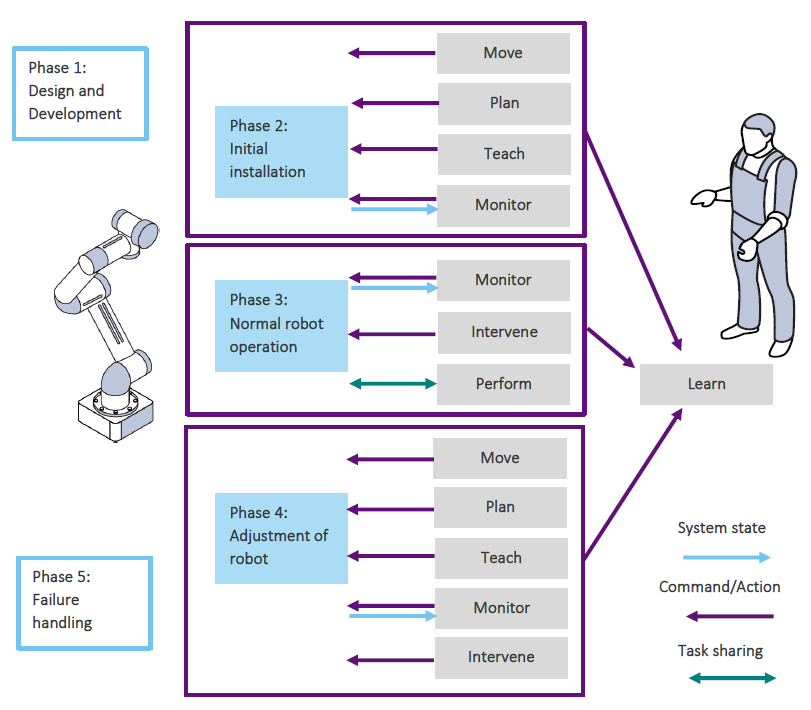}
  \caption{Model on which tasks are mostly relevant in which phase}
  \label{fig:taskphasemodel}
\end{figure}

The participants in the above-mentioned user research have been asked several times to mention the skills that the user needs to perform during the usage of the robot. The skills necessary for interaction with the user in Phase 3 are five times “None” and then “Monitoring”, “Communication”, “human interaction task”, “use interface”, “Assessing outcomes of tasks”, and “Transport the robot”. The task model from Sheridan is applied, which showed that the human is in nearly all phases quite active (least in phase 3) and that “Plan” tasks are dominant in phase 1 and phase 4, “Teach” tasks are dominant in Phase 2, “Monitor/Perform” is dominant in Phase 4, “Intervene” in Phase 5, and “Learn” is underrepresented. Within the analysis of the most important user groups, the question is asked again in a different way, the most important information appears to be “identify current state” (23 mentions in Figure \ref{fig:communicationneeds}), followed by “identify past state” (9 mentions) and “others” (9 mentions). These findings are summarized in Figure \ref{fig:taskphasemodel}, which displays the most relevant tasks for each phase and the interaction types.

\subsection{Limitations}

Due to the low response rate of the questionnaires from the side of companies, these results can only be considered preliminary. The student groups that have been interviewed as well were very informative, but it could be argued that older participants would have already had more project experience and thus would not neglect the needs of the users. What is indeed hard to measure is the high amount of people that did not want to fill out a questionnaire because they did not feel that their robotic project involved human-robot interaction (although the researcher did not share this impression based on their project presentations).
 
\subsection{Conclusion}
In general, it can be said and proved by the results from this investigation, that the human-robot interaction design is neglected by the development teams, who are majority male and computer-scientists. It is important to waken this sensibility in order to be able to develop robot systems that can be successful on the market instead of having systems that can only be used by people with a rare special background.

Neglecting the end-user within the design process is especially problematic as the identified highest priority user groups do not have a high technology affinity and are partly conservative. There have been two direct quotes in which groups mentioned that their robotic system is causing the users to be afraid of losing their jobs because they think that they are replaced by robots. Designing a highly complex system for this kind of user group will be very challenging without the proper interaction design knowledge.

\begin{acks}
The research presented here has received funding within the RobMoSys project H2020-EU.2.1.1. and ID 732410.
\end{acks}

\bibliographystyle{ACM-Reference-Format}
\bibliography{sample-base}
\end{document}